\pdfoutput=1
\documentclass[11pt]{article}

\usepackage[table]{xcolor}  

\usepackage[final]{acl}

\usepackage{times}
\usepackage{latexsym}
\usepackage{tabularx} 

\usepackage[T1]{fontenc}

\usepackage[utf8]{inputenc}

\usepackage{microtype}

\usepackage{inconsolata}
\usepackage{amsmath}
\usepackage{titlesec}

\usepackage{graphicx}

\title{Do Large Language Models Align with\\Core Mental Health Counseling Competencies?}

\author{
 \textbf{Viet Cuong Nguyen\textsuperscript{1}},
 \textbf{Mohammad Taher\textsuperscript{1}},
 \textbf{Dongwan Hong\textsuperscript{1}},
  \textbf{Vinicius Konkolics Possobom\textsuperscript{1}},\\
 \textbf{Vibha Thirunellayi Gopalakrishnan\textsuperscript{1}},
  \textbf{Ekta Raj\textsuperscript{1}},
 \textbf{Zihang Li\textsuperscript{2}},
 \textbf{Heather Jamie Soled\textsuperscript{3}},\\
 \textbf{Michael L. Birnbaum\textsuperscript{4}},
 \textbf{Srijan Kumar\textsuperscript{1}},
 \textbf{Munmun De Choudhury\textsuperscript{1}},
\\
\\
 \textsuperscript{1}Georgia Institute of Technology,
 \textsuperscript{2}Hofstra University,
 \textsuperscript{3}Tower Health,
 \textsuperscript{4}Columbia University,
\\
 \small{
   \textbf{Correspondence:} \href{johnny.nguyen@gatech.edu}{johnny.nguyen@gatech.edu}
 }
}

\begin{document}
\maketitle

\titlespacing*{\section}{0pt}{1ex plus 0.2ex minus 0.2ex}{1ex plus 0.2ex minus 0.2ex}

\begin{abstract}
The rapid evolution of Large Language Models (LLMs) presents a promising solution to the global shortage of mental health professionals. However, their alignment with essential counseling competencies remains underexplored. We introduce CounselingBench, a novel NCMHCE-based benchmark evaluating 22 general-purpose and medical-finetuned LLMs across five key competencies. While frontier models surpass minimum aptitude thresholds, they fall short of expert-level performance, excelling in Intake, Assessment \& Diagnosis but struggling with Core Counseling Attributes and Professional Practice \& Ethics. Surprisingly, medical LLMs do not outperform generalist models in accuracy, though they provide slightly better justifications while making more context-related errors. These findings highlight the challenges of developing AI for mental health counseling, particularly in competencies requiring empathy and nuanced reasoning. Our results underscore the need for specialized, fine-tuned models aligned with core mental health counseling competencies and supported by human oversight before real-world deployment. Code and data associated with this manuscript can be found at: \url{https://github.com/cuongnguyenx/CounselingBench}
\end{abstract}

\begin{table*}[t]
\footnotesize
\setlength{\tabcolsep}{2.5pt}
\begin{tabular}{p{3.0cm}p{4.0cm}p{8cm}c}
\hline
\textbf{Competency} & \textbf{Definition} & \textbf{Example Question} & \textbf{\#} \\
\hline
\textbf{Counseling Skills \& Interventions} (CS\&I) & 
Counselors' knowledge, skills, and abilities to conduct effective counseling. & 
\textbf{Q:} How would you approach the client's expressed fears concerning drinking?
\newline\textbf{A:} Your ethical responsibility is to advocate for this as an advancement of client care. & 
599 \\
\hline
\textbf{Intake, Assessment, \& Diagnosis} (IA\&D) & 
Counselors' knowledge, skills, and abilities to effectively conduct client intake, assessment, and diagnosis. & 
\textbf{Q:} Which assessment tool would you use to conduct a comprehensive evaluation of the client's current cognitive and emotional functioning?
\newline\textbf{A:} Beck Depression Inventory (BDI) & 
460 \\
\hline
\textbf{Professional Practice \& Ethics} (PP\&E) & 
Counselors' knowledge, skills, and abilities related to maintaining proper administrative and clinical protocols. & 
\textbf{Q:} Of the following, which is an ethically appropriate action to take to address the issue associated with your treatment facility's closing in six months?
\newline\textbf{A:} You express your advocacy for affordable treatment by writing an editorial for the newspaper. & 
274 \\
\hline
\textbf{Treatment Planning} (TP) & 
Counselors' knowledge, skills, and abilities to develop an effective course of treatment. & 
\textbf{Q:} You feel this client would benefit from additional professional help. To whom would you refer the client for help with her depression and sleep issues?
\newline\textbf{A:} Psychiatrist & 
253 \\
\hline
\textbf{Core Counseling Attributes} (CCA) & 
Behaviors, traits, and dispositions of effective counselors. & 
\textbf{Q:} You do not share the same religious views as the client. Which of the following would be the most helpful approach when considering the impact of your own beliefs in counseling?
\newline\textbf{A:} You continuously assess how your beliefs may affect the counseling process & 
23 \\
\hline
\end{tabular}
\caption{Description of key counseling competencies identified by the NBCC, along with example questions for each competency in CounselingBench}
\label{tab:m1}
\end{table*}

\section{Introduction}
Despite the critical importance of mental healthcare for individual and societal well-being, a significant global accessibility crisis continues to exist. Even in highly developed nations like the United States, access to adequate mental health services remains alarmingly insufficient. Current estimates indicate that more than half of the U.S. population resides in designated Mental Health Professional Shortage Areas (MHPSAs), regions where the number of mental health professionals falls short of meeting the population's needs~\cite{heisler2013mental}. This shortage poses a major public health challenge~\cite{heisler2013mental}, as it hinders timely intervention, contributes to untreated mental health conditions, and exacerbates disparities in care.

Language forms the foundation of mental healthcare, underpinning all interactions and interventions between patients and care providers. Recent advances in Large Language Models (LLMs) offer significant potential to alleviate the aforementioned global shortage in mental healthcare, attributed to their state-of-the-art performance in diverse natural language understanding tasks without additional fine-tuning. In fact, numerous research and commercial efforts have been directed at building LLM-based therapists and counselors to meet people's mental health needs~\cite{lai2023psy}. Increasing numbers of people are also appropriating general-purpose LLMs to find support and advice that may not be available through conventional means~\cite{de2023benefits}.

However, mental health is a fundamentally ``human'' experience, and addressing its challenges requires a nuanced blend of empathy, cultural sensitivity, and clinical expertise. Effective mental health care necessitates a range of competencies that are sensitive to the myriad and often complex manifestations of individuals' mental health journeys~\cite{clasen2003development}. These competencies include conducting thorough psychological assessments, which involve interpreting both verbal and non-verbal cues, understanding patients' unique life contexts, and identifying subtle signs of distress or underlying conditions~\cite{hoge2005workforce}. Additionally, clinicians must develop personalized treatment plans that account for each patient's background, personal history, and presenting symptoms~\cite{banikiotes1977training}. This process is not merely about applying standard diagnostic criteria; it involves setting treatment goals, choosing interventions that align with patients' values, and suggesting strategies based on patients' ongoing responses to treatment.

Furthermore, mental illnesses exhibit significant clinical heterogeneity~\cite{wardenaar2013diagnostic} -- not only are the conditions themselves diverse, encompassing everything from mood and anxiety disorders to complex cases of psychosis or personality disorders, but the symptoms and lived experiences manifest in deeply subjective ways. Effective mental healthcare, therefore, demands a flexible, patient-centered approach that can accommodate this diversity~\cite{zangeneh2019culture}. This raises critical questions about whether LLMs, despite their impressive capabilities in natural language understanding, can truly replicate the intricate, human-centered nature of mental health counseling. In this paper, we therefore seek to explore whether LLMs are equipped to demonstrate these core mental health counseling competencies. 

We present CounselingBench, a new benchmark designed to evaluate LLM performance in the context of key mental health counseling competencies. CounselingBench is based on the National Clinical Mental Health Counseling Examination (NCMHCE), a U.S. licensing exam for mental health counselors that assesses five core competencies identified by a broad survey of professionals in the field. Through CounselingBench, we systematically assess how well LLMs can process and apply domain-specific knowledge from case studies to address questions that evaluate these key competencies. Our research aims to address the following questions:\\
    \textbf{RQ1}: Are large language models capable of successfully passing the NCMHCE?\\
    \textbf{RQ2}: How accurately can large language models respond to NCMHCE questions covering various mental health counseling competencies?\\
    \textbf{RQ3} How effectively can LLMs generate rationales for their answers to these competency questions?\\
\indent Our contributions are as follows:\\
\indent \(\bullet\) We provide a novel benchmark, CounselingBench, designed to assess LLM capabilities across key mental health counseling competencies. This benchmark is based on the National Clinical Mental Health Counseling Examination (NCMHCE), ensuring its relevance and alignment with professional standards in the field.  \\
\indent \(\bullet\) We conduct a comprehensive evaluation of various LLMs, including both general-purpose and medical-specialized models, using CounselingBench. This evaluation provides insights into the current capabilities and limitations of LLMs in mental health counseling tasks.\\
\indent \(\bullet\) We analyze the performance of LLMs across different competencies, identifying areas of strength and weakness. This analysis can guide future fine-tuning efforts for LLMs and development of LLM-enabled tools for mental health applications. \\
\indent \(\bullet\) We examine the reasoning capabilities of LLMs in formulating responses to mental health counseling questions, shedding light on their ability to synthesize contextual information and domain-specific knowledge to explain their decisions in high-stakes counseling scenarios .\\
\indent \(\bullet\) We compare the performance of medical LLMs with general-purpose LLMs, highlighting specific shortcomings in the current medical models regarding mental health counseling competencies and identifying areas for potential improvement.

\section{Related Works}
\subsection{LLMs for Mental Health}
Online psychological counseling has experienced significant growth, especially in the wake of the COVID-19 pandemic \cite{yurayat2023university}. This increased demand has spurred research into the application of large language models (LLMs) for various mental health services \cite{stade2024large, lawrence2024opportunities}. However, mental healthcare is inherently complex, as it involves nuanced aspects of empathy, emotional intelligence, and context-specific interaction \cite{fried2020systems}. This complexity is compounded by the diversity of evaluation methods—ranging from analyses of annotated social media posts \cite{lamichhane2023evaluationchatgptnlpbasedmental} to assessments based on clinical vignettes \cite{levkovich2023vignette}—which makes cross-study comparisons challenging.

Recent research increasingly focuses on tailoring LLMs towards specific therapeutic interventions. For example, the Chain-of-interaction framework \cite{han2024chain} leverages dyadic contexts to enrich LLM understanding of psychiatric behaviors. This approach is designed to capture the iterative, interactive nature of counselor-client exchanges and potentially enhance the empathy and engagement critical to effective motivational interviewing. In parallel, efforts in Cognitive Behavior Therapy (CBT) have also emerged. CBT-BENCH \cite{zhang2024cbt} provides a structured evaluation of LLM capabilities in assisting CBT sessions, building upon previous work such \cite{chen2023empowering} which investigates the detection of cognitive distortions through diagnostic thought prompting. These advances demonstrate the growing trend of adapting LLMs to domain-specific therapeutic tasks, highlighting both their potential and the need for standardized, clinically grounded evaluation frameworks across multiple mental health intervention techniques and competencies.

Despite the promise shown by these specialized approaches, existing evaluation metrics often only moderately correlate with human clinical judgment, even though LLMs generally offer better explainability than traditional supervised models \cite{yang2024mentallama}. In contrast, our work introduces the first large-scale, mental health-specific evaluation benchmark that aligns with the competencies required of aspiring licensed mental health counselors, thereby aiming to bridge the gap between automated assessment and clinical practice.

\subsection{Competency Evaluation of LLMs}
The rapid evolution of LLMs has led to an extensive body of work evaluating their performance across a broad spectrum of tasks. General benchmarks have focused on natural language understanding, mathematical reasoning, coding, and even social science knowledge \cite{hendrycks2020measuring}. For instance, some benchmarks assess LLM performance on graduate-level academic information \cite{rein2023gpqa}, while others test mathematical reasoning abilities using grade-school problems \cite{cobbe2021training} or natural science word problems from the ARC dataset \cite{hu2024routerbench}.

In addition to these general assessments, several domain-specific benchmarks have been developed. The MedQA benchmark \cite{jin2021disease} evaluates LLMs' ability to process and respond to medical literature based on exam-style questions, and LegalBench \cite{legalBench2023} assesses legal reasoning skills. Building on this paradigm, our work extends the evaluation landscape by proposing the first comprehensive benchmark for assessing LLM proficiency in mental health counseling. This benchmark is designed not only to measure technical performance but also to capture the clinical and empathetic subtleties essential for real-world mental health practice, paving the way for more human-centered and clinically valid applications of LLM technology.

\section{Curating CounselingBench}
We collected the case study details (including patient demographic, mental status examination, presenting problem, etc.), questions, their associated answers and expert-generated rationale from National Clinical Mental Health Counseling Examination (NCMHCE) questions. The exam seeks to assess the proficiency of individuals seeking to become licensed clinical mental health counselors across five key mental health counseling competencies as detailed in Table \ref{tab:m1}. These competencies are derived from a national job analysis involving over 16,000 credentialed counselors, which identified empirically-validated work behaviors that are considered most relevant for effective counseling practice \cite{NBCC2023Outline}. NCMHCE questions and associate details listed above are collected from mock exams which are accessible online for public usage. Details about the sources that we collected data from are described in Appendix Table \ref{tab:a1}\\
\indent Overall, we collected a total of 1612 unique questions across 138 case studies that constitutes CounselingBench. CounselingBench contains cases whose subject exhibit diverse ethnic and cultural background in addition to presentation of mental health conditions. For instance, there exists considerable number of case studies whose patients are from ethnic minority backgrounds such as Black (20 case studies, 241 questions), Hispanic (13 case studies, 151 questions), Multiracial (4 case studies, 48 questions), Asian (4 case studies, 46 questions), Native American (2 case studies, 24 questions). In addition, there also exist several case studies in CounselingBench where the patient was not born in the United States where cultural competency is essential to correctly answer associated questions. CounselingBench also examines diverse mental health condition, with 51 unique conditions found among questions. Many such conditions have low prevalence in the real world, such as cyclothymic disorder (0.5-1\% of adults, 13 questons) or reactive attachment disorder of childhood (1\% of children, 13 questions). Table \ref{tab:aF} provides the full distribution of mental health conditions within CounselingBench

A full example question, along with the case study context, can be found in Appendix Table \ref{tab:a2}. To comply with fair use law, we adapt the procedure used in Jin et al. 2020 \cite{jin2021disease} and shuffle the order of answer options (while keeping track with the correct answer). Given that each of the questions are designed to uniquely assess candidates' abilities in one specific competency, we manually annotate all 1612 questions to provide a complete expert-derived question-competency mapping for downstream analysis. Two annotators, both of whom are collaborators on the paper and are medical doctors specializing in psychiatry, independently annotate all questions for one of five competencies as specified in Table \ref{tab:m1}. For each question, they were instructed to carefully read through the case study details, question statement and candidate answers in its entirety and then select the counseling competency that best described the competency which the question aims to test based on the NCMHCE Content Outline \cite{NBCC2023Outline}. This yielded 845 questions where the 2 annotators were in agreement about the competency annotation and 767 questions where they were in disagreement. For questions where the 2 annotators' labels disagreed, we invited a third annotator who is a experienced licensed mental health counselor (LMHC) based in the U.S. to provide the tiebreaking vote. The number of questions determined by experts annotators to reflect each of the five core mental health-related competencies can be found in Table \ref{tab:m1} 

\section{Methodology}
\subsection{Model Selection}
Our study encompasses 13 open-source medical models, selected for their outstanding performance across various biomedical NLP tasks. These models represent seven distinct finetuning architectures: BioMedGPT~\cite{zhang2023biomedgpt}, Asclepius~\cite{kweon2023publicly}, Meditron~\cite{chen2023meditron}, MentaLlama~\cite{yang2024mentallama}, ClinicalCamel~\cite{toma2023clinical}, Med42~\cite{christophe2024med42}, and OpenBioLLM~\cite{ankitpal2024openbiollm}. 
To provide a comprehensive analysis, we also include the corresponding un-finetuned base models, henceforth referred to as \textit{generalist models}, of the aforementioned medical models, adding 10 more models to our study. Both sets of medical models and their generalist counterparts were chosen to represent a wide range of model sizes. Finally, for benchmarking against state-of-the-art proprietary systems, we include gpt-4o-2024-08-06 in our comparison \cite{bubeck2023sparks, achiam2023gpt} due to its top performance on LLM leaderboards such as Chatbot Arena \cite{chiang2024chatbot}. Detailed information regarding model parameters and training data sources can be found in Appendix Table \ref{tab:a3}.

\subsection{Inference}
We formally define a question in the CounselingBench dataset as \(\mathcal{Q} = (\mathcal{C}, \mathcal{A}_{cand}, \mathcal{A}_{corr})\), where \(\mathcal{C}\) is the context (including question text and patient demographics), \(\mathcal{A}_{cand}\) are candidate answers, and \(\mathcal{A}_{corr}\) is the correct answer. \(\mathcal{Q}^h\) is an abridged version of \(\mathcal{Q}\) without \(A_{corr}\). To elicit answers from a large language model \(\mathcal{M}\) given \(\mathcal{Q}\), we use the following prompting strategies:
\begin{itemize}
    \item \textit{Zero-shot (ZS)}: \(\mathcal{M}\) answers \(\mathcal{Q}^h\) without task-specific training, using the prompt template in Appendix Table \ref{tab:a4}.  
    \item \textit{Few-shot (FS)}: \(\mathcal{M}\) answers \(\mathcal{Q}^h\) after seeing demonstrative questions \(\mathcal{Q}_1, ..., \mathcal{Q}_n\) \cite{brown2020language}, using the prompt template in Appendix Table \ref{tab:a5}. We use $n = 3$ random demonstrative examples.
    \item \textit{Chain-of-thought (CoT)}: We augment each few-shot example with a step-by-step explanation \(\mathcal{R}\) towards the correct answer. The input includes \((\mathcal{Q}_1, ..., \mathcal{Q}_n), (\mathcal{R}_1, ..., \mathcal{R}_n)\) and \(\mathcal{Q}^h\), using the prompt template in Appendix Table \ref{tab:a6}. We use the intermediary reasoning chains for comparison in RQ3.
\end{itemize}

We also test self-consistency (SC) decoding \cite{wang2022self} on all prompting strategies. While originally applied to CoT prompting, we extend SC to zero-shot and few-shot prompting by performing 5 samplings each at temperature t \(\in [0.2, 0.4, 0.6, 0.8, 1]\) and taking the majority label across all 25 samplings.

\section{RQ 1: Can LLMs pass the NCMHCE}
\subsection{Method}
To assess LLMs' performance on CounselingBench, we calculate overall accuracy on all 1612 questions. Given varying real-world NCMHCE passing scores, we use an average threshold of 63\% accuracy based on previous reported NCMHCE passing scores. To compare medical and generalist models, we employ paired t-tests between each medical model and its unfinetuned generalist counterpart.
\subsection{Results}
Table 2 shows the overall performance on the CounselingBench for all models tested across different prompting and decoding settings.
\definecolor{LightCyan2}{RGB}{224, 255, 255}
\begin{table}[t]
\centering
\footnotesize
\setlength{\tabcolsep}{1.5pt}
\begin{tabular}{lccccc}
\hline
\textbf{Model} & \textbf{ZS} & \textbf{\begin{tabular}[c]{@{}c@{}}ZS +\\SC\end{tabular}} & \textbf{FS} & \textbf{\begin{tabular}[c]{@{}c@{}}FS +\\SC\end{tabular}} & \textbf{\begin{tabular}[c]{@{}c@{}}FS +\\COT\end{tabular}} \\
\hline
\multicolumn{6}{l}{\textbf{\textit{Generalist Models}}} \\
\rowcolor{LightCyan2}
Llama-2-7B & .408 & .357 & .326 & .37 & .335 \\
Llama-2-7B-in & .432 & .444 & .451 & .461 & .412 \\
\rowcolor{LightCyan2}
Llama-3-8B-in & .622 & .646 & .643 & .654 & .595 \\
Llama-2-13B & .45 & .437 & .423 & .464 & .406 \\
\rowcolor{LightCyan2}
Llama-2-13B-in & .526 & .525 & .529 & .543 & .493 \\
Llama-2-70B & .596 & .575 & .59 & .644 & .241 \\
\rowcolor{LightCyan2}
Llama-2-70B-in & .616 & .616 & .631 & .644 & .431 \\
Llama-3-70B-in & \textbf{.717} & \textbf{.731} & \textbf{.734} & \textbf{.739} & \textbf{.71} \\
\hline
\multicolumn{6}{l}{\textbf{\textit{Medical Models}}} \\
\rowcolor{LightCyan2}
Meditron-7B & .258 & .293 & .244 & .318 & .066 \\
Asclepius-7B & .28 & .307 & .233 & .294 & .279 \\
\rowcolor{LightCyan2}
BioMedGPT-LM-7B & .409 & .381 & .386 & .432 & .105 \\
\rowcolor{LightCyan2}
Asclepius-Llama3-8B & .339 & .359 & .363 & .386 & .269 \\ 
OpenBioLLM-8B & .565 & .585 & .583 & .607 & .533 \\
\rowcolor{LightCyan2}
Llama3-Med42-8B & .639 & .638 & .643 & .654 & .572 \\
\rowcolor{LightCyan2}
Asclepius-13B & .338 & .381 & .32 & .38 & .257 \\
MentaLLaMA-13B-in & .452 & .461 & .485 & .488 & .373 \\
\rowcolor{LightCyan2}
ClinicalCamel-70B & .619 & .679 & .657 & .698 & .398 \\
Med42-70B & .68 & .679 & .57 & .687 & .635 \\
\rowcolor{LightCyan2}
Meditron-70B & .557 & .551 & .587 & .633 & .315 \\
Llama3-Med42-70B & .688 & .696 & .704 & .705 & .623 \\
\rowcolor{LightCyan2}
OpenBioLLM-70B & .698 & .725 & .716 & .734 & .686 \\
\hline
\multicolumn{6}{l}{\textbf{\textit{Proprietary Models}}} \\
\rowcolor{LightCyan2}
gpt4o & .78 & .765 & .723 & .748 & .767 \\
\hline
\end{tabular}
\caption{Accuracy of tested LLMs on CounselingBench across different prompting and decoding settings}
\label{tab:m2}
\end{table}

\subsection{Zero-shot LLMs can pass the NCMHCE}
We found that frontier LLMs (in September 2024) with zero-shot prompting are able to perform on NCMHCE at a level which exceeds the pass threshold of \(63\%\) accuracy as defined above. These passing models are primarily larger in size (5 out of 6 models have more than 70B parameters) and are all instruction-tuned (whether on a general instruction fine-tuning dataset or one specific to the biomedical domain). Not surprisingly, gpt4o is the best performing model among all the tested models, achieving a zero-shot accuracy of 0.78. However, it is also notable that the best open-source model (Llama3-70B-it) only performed slightly worse (8.8\% reduction in zero-shot accuracy) than gpt4o, despite being significantly smaller in parameter size. We also note that while there is a large gap in performance between the smallest and largest version of instruction-tuned Llama2 (0.432 vs 0.616, 42.6\% gain), that gap has substantially decreased between Llama3-8B-it and Llama3-70B-it (0.622 vs 0.717, 15.3\% gain). This suggests significant improvements in Llama3's model architecture and data curation processes compared to Llama2, and that current and future smaller-scale models are becoming increasingly viable for mental health counseling tasks while maintaining a lower computational footprint. On the flipside, performance trends on the CounselingBench also imply that performance gains from scaling up model sizes will diminish, or even disappear despite future advancements in model architecture, training data curation and procedure. This matches with observations from \citet{anwar2024foundational}, \citet{mckenzie2023inverse} and \citet{zhou2024larger} on the potential limits of LLM scaling laws. Finally, we note that while the performance of frontier LLMs exceed that of the minimum passing level, it remains substantially lower than that expert-level human performance, which we set at 90\% based on expert-level human scores on biomedical QA benchmarks such as MedMCQA and MedQA \cite{lievin2024can}
\subsection{Medical models underperformed generalist models across all settings accuracy-wise}
Most surprisingly, we notice that from Table 2 that a supermajority of generalist models seem to consistently outperform their medical fine-tuned counterparts (10 out of 13 pairs) under zero-shot setting, with an average difference of 4.2 percentage points between generalist-medical model pairs in zero-shot accuracy (maximum = 0.15pp, minimum = -0.084pp). We deploy paired $t$-test to assess significant differences between generalist and medical models' zero-shot accuracy on CounselingBench, and find a significant difference between these two distributions \((t=2.939, p=0.013)\). This suggests a systematic underperformance of medical LLMs on mental health counseling-related questions compared to their un-finetuned counterparts. We perform a more fine-grained evaluation of this underperformance in the next research question. 
\section{RQ 2: How do LLMs perform across different mental health competencies}
\subsection{Method}
To achieve a more nuanced, competency-centric assessment of model performance, we disaggregate the accuracy metrics in Table \ref{tab:m2} into five competency-specific accuracies. This approach: (1) evaluates model capabilities across key counseling domains, reflecting diverse patient needs, and (2) elucidates factors contributing to performance discrepancies between medical-specialist and generalist model pairs.

\subsection{LLMs are better at treatment planning, worse at counseling skills,  interventions}
Table 3 represents zero-shot accuracies of the 5 mental health counseling competencies across all tested LLMs. There are small yet significant variations across different competencies. Across all models, we found that their performance on different competencies (based on zero-shot accuracy) can be sorted as Treatment Planning = Intake, Assessment \& Diagnosis \(>^{*}\) professional practice \& ethics = core counseling attributes = counseling skills and interventions, where \(>\) indicates a statistically significant positive difference between zero-shot accuracy across competencies, with the number of stars in the superscript representing the p-value of the paired t-test\footnote{* p < 0.05, ** p < 0.01, *** p < 0.001}. Such variations remain even under other inference settings (such as few-shot). The observed variations in accuracy across mental health counseling competencies might be attributed to the nature of the tasks and the inherent strengths and limitations of LLMs. Competencies like \textit{Treatment Planning} and \textit{Intake, Assessment \& Diagnosis} are more procedural and well-represented in model training data, allowing LLMs to perform better here. In contrast, questions on competencies such as \textit{Counseling Skills and Interventions} and \textit{Core Counseling Attributes} require more nuanced social, emotional intelligence and cultural competency. These are areas where LLMs have been shown to struggle in even with careful prompting \cite{zhou2024real, chiu2024computational}. 
\subsection{Medical models significantly underperform their generalist counterparts across most competencies}
Comparing the cross-competency performance (measured with zero-shot accuracy) between generalist models and medical models with paired $t$-test, we  find that generalist models significantly overperform their paired medical counterparts across 4 out of 5 key mental health-related competencies: Intake, Asssessment \& Diagnosis ($t = 3.15, p = 0.027$), Treatment Planning ($t = 2.52, p = 0.035$), Counseling Skills \& Interventions ($t = 3.61, p = 0.024$), and Professional Practice \& Ethics ($t = 2.5, p = 0.035$). Among open-source models, no one model has the highest performance across all five competencies. OpenBioLLM-70B achieves the highest performance in treatment planning, whereas Llama3-70B-in achieves the highest performance in the remaining four out of five competencies (even outperforming gpt4o in the ``core counseling attribute" competency). These discrepancies in model performance persist even when alternative prompting techniques and decoding settings (such as few-shot, few-shot with self-consistency, etc.) are applied, as seen in Appendix Tables \ref{tab:a7} and \ref{tab:a8}. These observed differences might stem from variations in the training data and the patterns each LLM has learned. Medical LLMs are primarily fine-tuned on biomedical research papers and medical question answering datasets, which may give them increased performance on the ``treatment planning" competency. However, this seems to be at the expense of their retention of knowledge patterns required for the 4 other mental health competencies. In contrast, generalist models, trained and instruction-tuned on a broader range of data, possess and leverage more knowledge patterns that are needed for non-clinical mental health competencies such as counseling skills \& interventions in diverse settings. These variations highlight the challenges of developing AI systems for diverse mental health counseling skills, where the need for both emotional intelligence and clinical precision is paramount.
\begin{table}[!ht]
\footnotesize
    \centering
    \setlength{\tabcolsep}{3pt}
    \begin{tabular}{llllll}
    \hline
    model & IA\&D & TP & CS\&I & PP\&E & CCA \\ \hline
    \multicolumn{6}{l}{\textit{\textbf{Generalist Models}}} \\ 
    \rowcolor{LightCyan2}
    Llama2-7B & 0.433 & 0.418 & 0.390 & 0.402 & 0.392 \\ 
    Llama-2-13B & 0.465 & 0.469 & 0.452 & 0.446 & 0.575 \\ 
    \rowcolor{LightCyan2}
    Llama-2-70B & 0.628 & 0.628 & 0.589 & 0.569 & 0.479 \\ 
    Llama-2-7B-in & 0.449 & 0.473 & 0.419 & 0.427 & 0.217 \\ 
    \rowcolor{LightCyan2}
    Llama2-13B-in & 0.531 & 0.553 & 0.504 & 0.539 & 0.474 \\ 
    Llama-2-70B-in & 0.647 & 0.621 & 0.598 & 0.619 & 0.439 \\ 
    \rowcolor{LightCyan2}
    Llama3-8B-in & 0.641 & 0.629 & 0.640 & 0.579 & 0.603 \\ 
    Llama-3-70B-in & \textbf{0.730} & 0.689 & \textbf{0.741} & \textbf{0.699} & \textbf{0.650} \\ \hline
    \multicolumn{6}{l}{\textit{\textbf{Medical Models}}} \\
    BioMedGPT-7B & 0.436 & 0.437 & 0.362 & 0.426 & 0.487 \\ 
    \rowcolor{LightCyan2}
    OpenBioLLM-8B& 0.556 & 0.623 & 0.568 & 0.530 & 0.444 \\ 
    MentaLLaMA-13B & 0.473 & 0.480 & 0.427 & 0.448 & 0.576 \\ 
    \rowcolor{LightCyan2}
    Llama3-Med42-8B & 0.623 & 0.634 & 0.648 & 0.624 & 0.568 \\
    ClinicalCamel-70B & 0.663 & 0.624 & 0.596 & 0.607 & 0.560 \\ 
    \rowcolor{LightCyan2}
    Asclepius-13B& 0.290 & 0.380 & 0.351 & 0.347 & 0.437 \\ 
    Llama3-Asclepius-8B & 0.341 & 0.371 & 0.372 & 0.299 & 0.300 \\ 
    \rowcolor{LightCyan2}
    Meditron-7B & 0.267 & 0.262 & 0.242 & 0.282 & 0.168 \\ 
    Meditron-70B & 0.576 & 0.593 & 0.564 & 0.515 & 0.396 \\
    \rowcolor{LightCyan2}
    Llama3-Med42-70B& 0.707 & 0.704 & 0.676 & 0.668 & 0.576 \\
    Med42-70B & 0.671 & 0.689 & 0.704 & 0.683 & 0.515 \\ 
    \rowcolor{LightCyan2}
    Asclepius-7B & 0.244 & 0.318 & 0.280 & 0.279 & 0.482 \\ 
    \rowcolor{LightCyan2}
    OpenBioLLM-70B & 0.707 & \textbf{0.708} & 0.707 & 0.693 & 0.556 \\  \hline
    \multicolumn{6}{l}{\textit{\textbf{Proprietary Models}}} \\ 
    \rowcolor{LightCyan2}
    gpt4o & 0.808 & 0.750 & 0.774 & 0.749 & 0.608 \\ \hline
    \end{tabular}
    \caption{Zero-shot accuracy of tested LLMs on CounselingBench, segmented across different counseling competencies and model types}
    \label{tab:m3}
\end{table}

\begin{figure}[h!]
    \centering
    \includegraphics[width=0.48\textwidth]{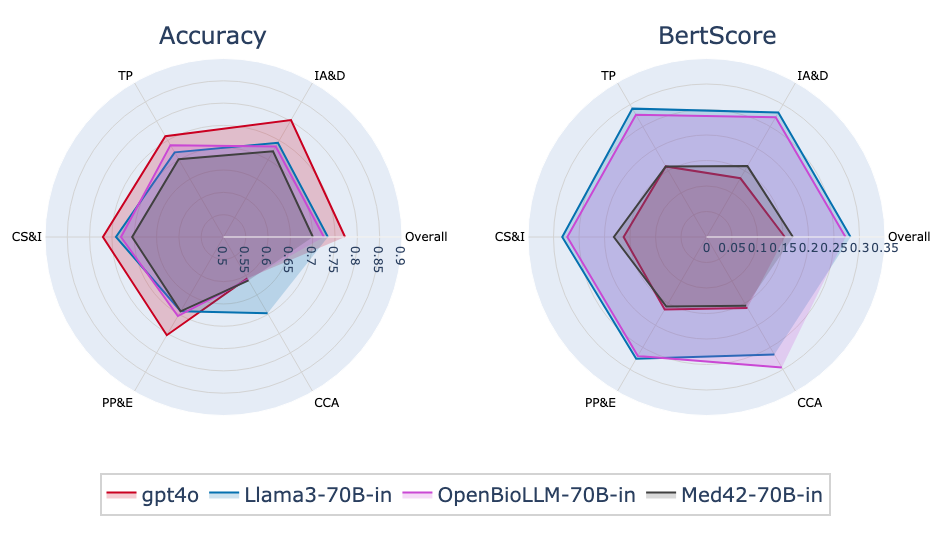} 
    \caption{Radar Chart indicating average zero-shot accuracy and BertScore of top-4 best performing models, across key counseling competencies}
    \label{fig:fig1}
\end{figure}

\section{RQ 3: How well do LLMs reason to derive their answers across different competencies?}

\begin{table*}[!ht]
\footnotesize
\centering
\setlength{\tabcolsep}{3.3pt}
\newcommand{\graycell}[1]{\cellcolor{gray!#1}}
\newcommand{\heatmap}[1]{%
  \ifdim#1pt<0.1pt
    \graycell{0}
  \else\ifdim#1pt<0.2pt
    \graycell{15}
  \else\ifdim#1pt<0.3pt
    \graycell{30}
  \else\ifdim#1pt<0.4pt
    \graycell{45}
  \else\ifdim#1pt<0.5pt
    \graycell{60}
  \else\ifdim#1pt<0.6pt
    \graycell{70}
  \else\ifdim#1pt<0.7pt
    \graycell{80}
  \else\ifdim#1pt<0.8pt
    \graycell{90}
  \else
    \graycell{100}
  \fi\fi\fi\fi\fi\fi\fi\fi
}
\begin{tabular}{lllllllllllll}
\hline
model & cosSim & bert & \(R_L\) & \(R_1\) & faith & \(info_{stp}\) & \(info_{chn}\) & mis. & al. & rep. \(\downarrow\) & gmr. & cons.\\
\hline
\textbf{Generalist Models} & & & & & & & & & & & &\\
\hline
Llama3-8B-in & \heatmap{0.740}0.740 & \heatmap{0.216}0.216 & \heatmap{0.221}0.221 & \heatmap{0.361}0.361 & \heatmap{0.738}0.738 & \heatmap{0.710}0.710 & \heatmap{0.876}0.876 & \heatmap{0.701}0.701 & \heatmap{0.807}0.807 & \heatmap{0.091}0.091 & \heatmap{0.884}0.884 & \heatmap{0.843}0.843 \\
Llama2-70B & \heatmap{0.256}0.256 & \heatmap{0.046}0.046 & \heatmap{0.053}0.053 & \heatmap{0.080}0.080 & \heatmap{0.547}0.547 & \heatmap{0.516}0.516 & \heatmap{0.642}0.642 & \heatmap{0.478}0.478 & \heatmap{0.586}0.586 & \heatmap{0.243}0.243 & \heatmap{0.687}0.687 & \heatmap{0.624}0.624 \\
Llama2-70B-in & \heatmap{0.531}0.531 & \heatmap{0.162}0.162 & \heatmap{0.170}0.170 & \heatmap{0.276}0.276 & \heatmap{0.586}0.586 & \heatmap{0.558}0.558 & \heatmap{0.733}0.733 & \heatmap{0.521}0.521 & \heatmap{0.630}0.630 & \heatmap{0.581}0.581 & \heatmap{0.825}0.825 & \heatmap{0.793}0.793 \\
Llama3-70B-in & \heatmap{0.769}\textbf{0.769} & \heatmap{0.282}\textbf{0.282} & \heatmap{0.255}\textbf{0.255} & \heatmap{0.402}\textbf{0.402} & \heatmap{0.685}0.685 & \heatmap{0.647}0.647 & \heatmap{0.831}0.831 & \heatmap{0.621}0.621 & \heatmap{0.841}0.841 & \heatmap{0.644}0.644 & \heatmap{0.937}0.937 & \heatmap{0.952}\textbf{0.952} \\
\hline
\textbf{Medical Models} & & & & & & & & & & & & \\
\hline 
Llama3-Med42-8B & \heatmap{0.662}0.662 & \heatmap{0.235}0.235 & \heatmap{0.220}0.220 & \heatmap{0.352}0.352 & \heatmap{0.699}0.699 & \heatmap{0.664}0.664 & \heatmap{0.840}0.840 & \heatmap{0.627}0.627 & \heatmap{0.728}0.728 & \heatmap{0.481}0.481 & \heatmap{0.897}0.897 & \heatmap{0.837}0.837 \\
OpenBioLLM-8B & \heatmap{0.739}0.739 & \heatmap{0.269}0.269 & \heatmap{0.242}0.242 & \heatmap{0.379}0.379 & \heatmap{0.719}0.719 & \heatmap{0.694}0.694 & \heatmap{0.874}0.874 & \heatmap{0.712}0.712 & \heatmap{0.844}0.844 & \heatmap{0.109}0.109 & \heatmap{0.970}0.970 & \heatmap{0.833}0.833 \\
meditron-70B & \heatmap{0.351}0.351 & \heatmap{0.072}0.072 & \heatmap{0.088}0.088 & \heatmap{0.139}0.139 & \heatmap{0.552}0.552 & \heatmap{0.522}0.522 & \heatmap{0.652}0.652 & \heatmap{0.480}0.480 & \heatmap{0.597}0.597 & \heatmap{0.280}0.280 & \heatmap{0.700}0.700 & \heatmap{0.637}0.637 \\
ClinicalCamel-70B & \heatmap{0.491}0.491 & \heatmap{0.130}0.130 & \heatmap{0.118}0.118 & \heatmap{0.187}0.187 & \heatmap{0.682}0.682 & \heatmap{0.650}0.650 & \heatmap{0.831}0.831 & \heatmap{0.603}0.603 & \heatmap{0.704}0.704 & \heatmap{0.488}0.488 & \heatmap{0.885}0.885 & \heatmap{0.680}0.680 \\
med42-70B & \heatmap{0.729}0.729 & \heatmap{0.250}0.250 & \heatmap{0.235}0.235 & \heatmap{0.368}0.368 & \heatmap{0.715}0.715 & \heatmap{0.686}0.686 & \heatmap{0.863}0.863 & \heatmap{0.674}0.674 & \heatmap{0.817}0.817 & \heatmap{0.210}0.210 & \heatmap{0.951}0.951 & \heatmap{0.863}0.863 \\
Llama3-Med42-70B & \heatmap{0.694}0.694 & \heatmap{0.170}0.170 & \heatmap{0.177}0.177 & \heatmap{0.322}0.322 & \heatmap{0.752}\textbf{0.752} & \heatmap{0.728}\textbf{0.728} & \heatmap{0.898}\textbf{0.898} & \heatmap{0.678}0.678 & \heatmap{0.770}0.770 & \heatmap{0.076}\textbf{0.076} & \heatmap{0.940}0.940 & \heatmap{0.427}0.427 \\
OpenBioLLM-70B & \heatmap{0.749}0.749 & \heatmap{0.273}0.273 & \heatmap{0.245}0.245 & \heatmap{0.388}0.388 & \heatmap{0.723}0.723 & \heatmap{0.700}0.700 & \heatmap{0.877}0.877 & \heatmap{0.722}\textbf{0.722} & \heatmap{0.846}\textbf{0.846} & \heatmap{0.109}0.109 & \heatmap{0.972}\textbf{0.972} & \heatmap{0.804}0.804 \\
\hline
\textbf{Proprietary Models} & & & & & & & & & & & & \\
\hline
gpt4o & \heatmap{0.615}0.615 & \heatmap{0.154}0.154 & \heatmap{0.160}0.160 & \heatmap{0.262}0.262 & \heatmap{0.730}0.730 & \heatmap{0.701}0.701 & \heatmap{0.873}0.873 & \heatmap{0.705}0.705 & \heatmap{0.831}0.831 & \heatmap{0.161}0.161 & \heatmap{0.968}0.968 & \heatmap{0.841}0.841  \\
\hline
\end{tabular}
\caption{Heatmap representing average reasoning quality metrics of high-performing LLMs. Down arrow \((\downarrow)\) indicates lower scores equals better reasoning quality in that metric. Acronyms in columns correspond to metrics as described in Appendix Table \ref{tab:a9}}
\label{tab:m4}
\end{table*}

\subsection{Method}
Beyond evaluating LLMs' accuracy in mental health counseling competencies, we assess their ability to generate high-quality reasoning chains based on the question and case context. This is crucial for future applications, as clear and coherent reasoning enhances LLMs' reliability in guiding therapeutic conversations, potentially improving outcomes for individuals seeking mental health support. Poor reasoning or misinterpretations in therapeutic contexts can compromise treatment and even worsen conditions such as anxiety, depression, or psychosis~\cite{obradovich2024opportunities, de2023benefits}. We evaluate reasoning quality using the intermediary chains produced during the chain-of-thought prompting process (Section 4.2). These reasoning chains, or \textit{candidate chains}, are assessed along key axes—alignment with expert reasoning, coherence, and informativeness—using both reference-free and reference-based metrics:
\begin{itemize}
    \item Reference-free metrics assess candidate chains without relying on reference answers, considering factors like semantic alignment, logical inference, and language coherence. We use ROSCOE, a suite of unsupervised evaluation metrics, due to its scalability and strong correlation with human judgments~\cite{golovneva2022roscoe} with subset of metrics as describe in Appendix Table \ref{tab:a9}.
    \item Reference-based metrics directly compare candidate chains to expert-generated \textit{reference chains}, using metrics such as ROUGE-1, ROUGE-L~\cite{lin2004rouge}, BERTScore~\cite{zhang2019bertscore}, and cosine similarity.
\end{itemize}
To better understand the reasoning errors made by LLMs, we annotate errors in a subset of responses from two top-performing models, OpenBioLLM-70B and Llama3-70B-in. Three expert annotators reviewed 100 questions where these models had incorrect answers, categorizing errors as \textit{logical}, \textit{knowledge-based}, or \textit{contextual}, following patterns identified in medical reasoning~\cite{singhal2023towards} (Appendix Table \ref{tab:a10}).

\subsection{Results}
Table 4 presents both reference-free and reference-based metrics derived from the reasoning chains of models with a zero-shot multiple-choice accuracy of at least 0.5. Consistent with trends in multiple-choice accuracy as seen in Table \ref{tab:m2}, we observe that larger and more current open-source models show substantial improvements in generating reasoning chains that align with gold-standard reasoning and exhibit high-quality characteristics over smaller and earlier models, with smaller gaps between different model sizes for the current ``generation" of open-source LLMs. For instance, we noticed that the average cosine similarity of Llama3-70B-in is 3.3\% and 80\% higher than Llama3-8B-in and Llama2-70B-in respectively. 
Unlike multiple-choice accuracy trends, generalist models do not exhibit significantly better reasoning compared to specialized medical models, or vice versa. However, we do find that while no single model excels across all reasoning metrics, medical models achieve the highest score across a majority of reasoning metrics (7 / 12 \(\approx\) 58\%) and all but one reference-free metrics. While propriety frontier models such as gpt4o excel in answering multiple-choice questions over open-source models, they surprisingly produce lower-quality reasoning chains on average compared to some smaller yet more-performant medical open-source counterparts in all reasoning metrics. Appendix Table \ref{tab:a11} gives an example of step-by-step reasoning chains on a sample CounselingBench question among top-performing LLMs (with respective to both accuracy and reasoning metrics).  These findings highlight the capacity of training high-quality domain-specific models for producing justifications that is more aligned with expert-level decision process, particularly in specialized areas like mental healthcare, despite their diminished ability to derive the correct answer compared to their generalist counterparts. We also note that frontier LLMs still underperform expert-level humans when it comes to producing high-quality reasoning chains to justify mental health counseling-related decisions, highlighting another shortcoming of existing frontier LLMs in the mental health domain (Appendix Table \ref{tab:a12}) 
\subsubsection{Analysis of reasoning errors}
We found that among 100 randomly sampled erroneous answers, Llama 3 70B made 33 logical errors, 21 context errors, and 46 knowledge errors in their reasoning, whereas OpenBioLLM-70B made 35 logical errors, 36 context errors and 29 knowledge errors. A statistical analysis using $\chi^2$ test reveals a significant difference in the distribution of errors made by these two models representative of generalist and medical LLMs, where generalist models are more likely to make knowledge-related errors during their reasoning process whereas medical models are more likely to make reading comprehension-related errors (\(\chi^2 = 17.12\), $p$ \(<\) 0.001). This suggests that while generalist models may struggle with domain-specific knowledge, specialized models face challenges in accurately interpreting context, underscoring a potential trade-off in reasoning capabilities during the fine-tuning process. 
\section{Conclusion and Future Work}
To effectively integrate LLMs into real-world mental health counseling, rigorous assessments of their alignment with core competencies are necessary. In this study, CounselingBench, a benchmark for evaluating LLMs in mental health counseling, was introduced. Both un-finetuned and fine-tuned medical models demonstrated strong capabilities across five key counseling competencies but showed varying performance levels. LLMs excelled in \textit{Intake, Assessment \& Diagnosis} but underperformed in \textit{Core Counseling Attributes} and \textit{Professional Practice \& Ethics}, which require greater subjectivity and sensitivity to individual patient contexts. Fine-tuning on biomedical data did not consistently improve performance, possibly due to limited representation of mental health-specific competencies in the training data. In some cases, these models even underperformed compared to generalist models. This indicates improving LLMs for mental health tasks may require more targeted fine-tuning with diverse, real-world counseling data. Future work should focus on developing specialized LLMs for counseling and adopt more task-specific evaluations to address the unique demands of each competency.
\section{Limitations}
\subsection{Multiple-choice Exams Cannot Fully Capture Proficiency in Counseling Competencies}
Evaluating LLM competency via multiple-choice exams may not fully capture the complexities of real-world mental health counseling. While multiple-choice formats offer a standardized way to measure knowledge, they often fail to fully assess critical aspects of mental health counseling such as empathy, adaptability, and nuanced decision-making, which are essential in clinical practice. Real-world counseling involves dynamic and context-sensitive interactions that go beyond selecting a correct answer from given options. As a result, models that perform well on these exams may not necessarily demonstrate the same level of competency in genuine therapeutic settings, where understanding the patient's unique background and responding to evolving emotional states are key factors \cite{ obradovich2024opportunities}. Thus, relying solely on multiple-choice accuracy may oversimplify the challenges of mental health counseling and provide an incomplete picture of LLM capabilities in this domain.
\subsection{Scale and Scope of CounselingBench}
Second, the scale and scope of the CounselingBench dataset present limitations. Given the size of the dataset, it may not adequately represent the diversity of situations encountered by mental health professionals. The current questions are based on mock exams for the NCMHCE, which may not encompass the full range of clinical presentations and treatment scenarios, especially for culturally specific or less common conditions \cite{hoge2005workforce, zangeneh2019culture}. In addition, since the NCMHCE is a US-based licensing exam, despite containing questions regarding patients from a wide range of background, it does not fully cover mental health counseling scenarios and best practices in other global regions such as the Global South, which has significant differences in cultural values, social norms, and access to mental health resources \cite{sue2022counseling}. Such conditions require tailored approaches to mental health counseling and intervention that are not adequately reflected by the exam's current content. Expanding the dataset to include a wider variety of case studies, question formats, and counseling competencies would enhance its representativeness and allow for a more comprehensive evaluation of LLM performance. Additionally, integrating open-ended and scenario-based questions could better assess the models' ability to engage in complex reasoning and provide contextually appropriate responses, thereby improving the generalizability of findings.
\section{Ethical Considerations}
We note that CounselingBench may be used in future research projects to make sweeping claims regarding LLMs outperforming human mental health counselors and could potentially replace them in real-world mental health counseling situations, similar to claims made in the biomedical domain\cite{drogt2024ethical}. Such claims. even with higher performance of LLMs on CounselingBench, would be premature and potentially harmful given the lack of longitudinal assessment for LLMs, and incomplete evaluation of its decision-making and execution process. Overreliance on AI systems could compromise patient care and contribute to the erosion of therapeutic alliance ~\cite{choudhury2024large, de2023benefits}. We strongly advise against using CounselingBench to argue for replacing human professionals; instead, it should be viewed as a tool for enhancing AI assistants in tasks supportive of human mental health professionals \cite{van2023global}. To ensure responsible development and application of AI in mental health, we believe that interdisciplinary collaboration among AI researchers, mental health professionals, ethicists, and policymakers is essential.
\section{Acknowledgements}
Viet Cuong Nguyen is partially supported by NSF \#2230692, NIH R01MH117172. This study is supported by the Microsoft Accelerating Foundation Models Research program. We acknowledge the assistance of Santiago Alvarez Lesmes and Theodore Vlavianos in labeling CounselingBench. We also thank the members of SOCWEB and CLAWS lab at Georgia Institute of Technology for giving feedback throughout the entire process of creating this manuscript.

\bibliography{acl_latex}


\appendix
\setcounter{table}{0}
\renewcommand{\thetable}{A\arabic{table}}
\section*{Appendix}  

\section{Additional Details on Experiments}
\textbf{Benchmark Evaluation} For all inference cases, we use Nucleus Sampling \cite{holtzman2019curious} with a probability threshold of 0.9 and a temperature of 0 to consistently generate the highest-probability next token. We generate only 1 token for non-CoT inference setups, and 500 tokens for CoT inference setups. The temperature for inference setups with self-consistency are described in Section 4.2 above. All experiments were conducted on a server with 5 A100 80GB GPUs. All models are implemented and initialized for inference with vLLM \cite{kwon2023efficient}, parallelized across 4 GPUs with \textit{gpu\_memory\_utilization} set to 0.5. We use scikit-learn to calculate accuracy results from LLM experiments \cite{pedregosa2011scikit}, and scipy to perform all statistical procedures \cite{virtanen2020scipy}. To calculate all ROSCOE metrics, we utilize the official code implementation\footnote{https://github.com/facebookresearch/ParlAI/} as described in \cite{golovneva2022roscoe}. We use the official BERTScore\footnote{https://pypi.org/project/bert-score/} and the rouge-score packages \footnote{https://pypi.org/project/rouge-score/} to calculate BERTScore and ROUGE scores respectively. 

\section{Additional Details on Dataset}
\subsection{Data Gathering Process}

The dataset was initially gathered from publicly available sources on NCMHCE case studies and questions as described in Appendix Table \ref{tab:a1}. Using Python and BeautifulSoup, we automated the extraction of structured information from a CSV file containing exam questions. The data extraction process involved splitting and organizing the raw content as seen in Appendix Table \ref{tab:a2} into meaningful sections.

\subsection{Data Contamination Detection}

To ensure the integrity of the collected data, we adopted a data contamination detection methodology based on the approach presented in \cite{hernandez2023measuring}. This process was designed to identify any overlapping or low-quality data that could impact the evaluation of the language model. The core of the analysis involved generating the second half of counseling questions based on the first half and comparing the generated results with the original human-written second half. The following steps outline the process:

\begin{enumerate}
    \item \textbf{Input and Generation} 
    
    For each data entry, the first half of a question was provided as input, and the language model was tasked with generating the second half. This simulates real-world scenarios where a partially completed question is given, and the model must generate the rest in a coherent and contextually appropriate manner.
    
    \item \textbf{Comparison between the Second Half} 
    
    Once the second half was generated, we compared it with the original, human-written second half. Two main techniques were used for this comparison:
    
    \begin{itemize}
        \item \textbf{Levenshtein Distance} 
        
        A string-based similarity measure that calculated the number of character-level changes needed to transform the generated second half into the original version. This helped identify cases of significant textual overlap or differences.
        
        \item \textbf{BERT-based Semantic Similarity} 
        
        A model-based similarity measure that evaluated how closely the generated text matched the original in terms of meaning. This ensured that even if the generated text was not an exact match, its semantic coherence could still be assessed.
    \end{itemize}
    
    \item \textbf{Contamination Check} 
    
    We set contamination thresholds for the BERT-based semantic similarity scores, following the guidelines from \cite{hernandez2023measuring}. A threshold of 0.9 was used to flag entries that were too similar to the original human-written content. Our analysis found that \textbf{0 entries} exceeded this 0.9 threshold, meaning no entries showed extreme similarity. However, \textbf{7 entries} exceeded the 0.8 threshold, which still indicated some level of overlap but did not qualify as data contamination. These entries were reviewed but ultimately deemed acceptable for inclusion in the dataset. After reviewing the flagged entries and ensuring that they did not represent significant data contamination, we concluded that the case studies and questions constituting CounselingBench are not likely to appear in the training data of LLMs included in our experiments. Therefore, the performance of LLMs on CounselingBench as shown above cannot be attributed to memorization but rather its innate ability to generalize based on its training data
\end{enumerate}

\section{Supplementary Information}

\begin{table}[ht]
\centering
\begin{tabularx}{\linewidth}{lXX}
\hline
Data Source        & \# Question & \# Case Studies \\ \hline
Mometrix.com       & 564         & 43              \\ 
Tests.com          & 509         & 51              \\ 
CounselingExam.com & 539         & 44              \\ \hline
\end{tabularx}
\caption{Data Sources constituting CounselingBench, number of questions and case studies per data source}
\label{tab:a1}
\end{table}

\begin{table*}[ht]
  \centering
  \label{tab:mh_prevalence}
  \begin{tabularx}{\textwidth}{|p{0.65\linewidth}|p{0.30\linewidth}|}
    \hline
    \textbf{MH Condition} & \textbf{Prevalence in CounselingBench} \\ \hline
    Reaction to severe stress, and adjustment disorders (F43) & 144 questions (8.9\%) \\ \hline
    Other anxiety disorders (F41) & 115 questions (7.1\%) \\ \hline
    Persistent mood [affective] disorders (F34) & 96 questions (6.0\%) \\ \hline
    Conduct disorders (F91) & 76 questions (4.7\%) \\ \hline
    Attention-deficit hyperactivity disorders (F90) & 69 questions (4.3\%) \\ \hline
    Eating disorders (F50) & 65 questions (4.0\%) \\ \hline
    Major depressive disorder, recurrent (F33) & 59 questions (3.7\%) \\ \hline
    Specific personality disorders (F60) & 53 questions (3.3\%) \\ \hline
    Alcohol related disorders (F10) & 52 questions (3.2\%) \\ \hline
    Bipolar disorder (F31) & 49 questions (3.0\%) \\ \hline
    Problems related to social environment (Z60) & 48 questions (3.0\%) \\ \hline
    Gender identity disorders (F64) & 47 questions (2.9\%) \\ \hline
    Pervasive developmental disorders (F84) & 38 questions (2.4\%) \\ \hline
    Other problems related to primary support group, including family circumstances (Z63) & 36 questions (2.2\%) \\ \hline
    Phobic anxiety disorders (F40) & 36 questions (2.2\%) \\ \hline
    Major depressive disorder, single episode (F32) & 23 questions (1.4\%) \\ \hline
    Occupant of heavy transport vehicle injured in collision with two- or three-wheeled motor vehicle (V62) & 13 questions (0.8\%) \\ \hline
    Schizophrenia (F20) & 13 questions (0.8\%) \\ \hline
    Somatoform disorders (F45) & 13 questions (0.8\%) \\ \hline
    Opioid related disorders (F11) & 13 questions (0.8\%) \\ \hline
    Disorders of social functioning with onset specific to childhood and adolescence (F94) & 13 questions (0.8\%) \\ \hline
    Problems related to other psychosocial circumstances (Z65) & 13 questions (0.8\%) \\ \hline
    Other symptoms and signs involving cognitive functions and awareness (R41) & 13 questions (0.8\%) \\ \hline
    Inhalant related disorders (F18) & 13 questions (0.8\%) \\ \hline
    Emotional disorders with onset specific to childhood (F93) & 12 questions (0.7\%) \\ \hline
    Problems related to upbringing (Z62) & 12 questions (0.7\%) \\ \hline
    Persons encountering health services for other counseling and medical advice, not elsewhere classified (Z71) & 10 questions (0.6\%) \\ \hline
    Dementia in other diseases classified elsewhere (F02) & 10 questions (0.6\%) \\ \hline
    Specific developmental disorders of scholastic skills (F81) & 10 questions (0.6\%) \\ \hline
  \end{tabularx}
    \caption{Prevalence of Mental Health Conditions in CounselingBench}
    \label{tab:aF}
\end{table*}

\begin{table*}[ht!]
\centering
\begin{tabularx}{\textwidth}{p{0.9\textwidth}}
\hline
\textbf{Patient Demographics} \\ 
\textit{Age}: 26 \\
\textit{Sex}: Male \\
\textit{Gender}: Male \\
\textit{Sexuality}: Heterosexual \\
\textit{Ethnicity}: Caucasian \\
\textit{Relationship Status}: Single \\
\textit{Counseling Setting}: Community Mental Health Agency \\
\textit{Type of Counseling}: Individual \\
\textit{Presenting Problem}: Hallucinations and Delusions \\
\textit{Diagnosis}: Schizophrenia 295.90 (F20.9) \\
\textbf{Mental Status Examination} \\
The client displays an angry affect, and his mood is irritable. His speech is disorganized and pressured. He is oriented to person, place, time, and situation. He reports audiovisual hallucinations, including seeing "the shadow man" and hearing voices others cannot hear. The client exhibits tangential and disconnected thinking. He is firm in his conviction that he is being poisoned and says he is exhausted from constantly trying to maintain vigilance. His insight and judgment are poor. He denies suicidal ideation, homicidal ideation, and command hallucinations. Symptoms began in his late teens but were misdiagnosed as bipolar disorder. \\
\textbf{Presenting Problem} \\
You are a counselor at an outpatient community mental health center serving clients with severe psychiatric disorders. A 26-year-old male, accompanied by his caseworker, presents for counseling due to symptoms of schizophrenia. The caseworker reports that the client was stable until he stopped taking his medication. He resides in assisted living, placed there after being discharged from the hospital last month. The client believes "the shadow man" is following him and poisoning his food. He has become more agitated, engaging in verbal altercations with other residents, and refuses medication due to side effects like restlessness and nervousness. The client is adamant about staying off medication and becomes angry when his caseworker mentions hospitalization. \\
\textbf{Question} \\
You administer the Scale for the Assessment of Positive Symptoms (SAPS) to determine the severity of which of the following? \\
\textbf{Answer Choices} \\
(A): Avolition \\
(B): Diminished speech \\
(C): Agitation \\
(D): Social withdrawal \\
\textbf{Correct Answer} \\
(C): Agitation \\
\textbf{Expert-generated Reasoning} \\
The SAPS measures the severity of positive symptoms in schizophrenia, including hallucinations, delusions, bizarre behavior, and thought disorders. Agitation is categorized under bizarre behavior, making (C) the correct answer. Avolition, diminished speech, and social withdrawal are negative symptoms. Avolition is a lack of goal-directed activity, diminished speech refers to reduced fluency, and social withdrawal indicates limited social interaction.
\\ \hline
\end{tabularx}
\caption{Example full question in CounselingBench regarding the \textit{Intake, Assessment \& Diagnosis} competency}
\label{tab:a2}
\end{table*}
\begin{table*}[ht!]
\footnotesize
\centering
\begin{tabularx}{\textwidth}{lXlXXXXX}
\hline
\textbf{Model Name}   & \textbf{Release Date}   & \textbf{Finetuning Data}                                                                                  & \textbf{Finetuned from} & \textbf{Number of Params} & \textbf{License} & \textbf{Reference} \\ \hline
\multicolumn{6}{l}{\textbf{GENERALIST MODELS}} \\
Llama                 & Feb 2023                & Unknown                                                                                                    & N/A                     & 7B, 13B    & Non-commercial              & \cite{touvron2023llama} \\ 
Llama-2               & Jul 2023                & Unknown                                                                                                    & N/A                     & 7B, 13B, 70B   & Llama 2 Community          & \cite{touvron2023llamaB} \\ 
Llama-2-in            & Jul 2023                & Unknown                                                                                                    & Llama-2              & 7B, 13B, 70B   & Llama 2 Community  & \cite{touvron2023llamaB} \\
Llama-3-in            & Apr 2024                & Unknown                                                                                                    & N/A                     & 8B, 70B     & Llama 3 Community   & \cite{dubey2024llama} \\ \hline
\multicolumn{6}{l}{\textbf{MEDICAL MODELS}} \\
BioMedGPT          & Aug 2023                & Biomedical Papers                                                                                          & Llama-2-in              & 7B            & Apache 2.0           & \cite{zhang2023biomedgpt} \\ 
Asclepius          & Sep 2023                & Synthetic Clinical Notes                                                                                   & Llama-2                 & 7B, 13B       &  CC-BY-NC-SA 4.0          & \cite{kweon2023publicly} \\ 
Asclepius         & Jun 2024                & Synthetic Clinical Notes                                                                                   & Llama-3-in              & 8B             &  CC-BY-NC-SA 4.0          & \cite{kweon2023publicly} \\ 
Med42-v2              & Aug 2024                & Medical Question-Answering Datasets                                                                        & Llama-3-in              & 8B, 70B    & Med42 Custom              & \cite{christophe2024med42} \\ 
OpenBioLLM            & May 2024                & Unknown                                                                                                    & Llama-3-in              & 8B, 70B    & Llama 3 Community             & \cite{ankitpal2024openbiollm} \\ 
Medalpaca             & May 2023                & \begin{tabular}[c]{@{}l@{}}Medical Flashcards,\\ Medical Question-\\Answering Datasets\end{tabular}           & Llama                   & 7B, 13B     & CC-BY-4.0           & \cite{han2023medalpaca} \\ 
ClinicalCamel         & Aug 2023                & \begin{tabular}[c]{@{}l@{}}Biomedical Papers,\\ Medical Question-\\Answering Datasets\end{tabular}            & Llama-2                 & 70B     & CC-BY-NC-4.0                & \cite{toma2023clinical} \\ 
Med42                 & Aug 2023                & Medical Question-Answering Datasets                                                                        & Llama-2                 & 70B        & Med42 Custom    & \cite{christophe2024med42} \\ 
Meditron          & Nov 2023                & \begin{tabular}[c]{@{}l@{}}Clinical Guidelines,\\ Biomedical Papers\end{tabular}                           & Llama-2                 & 7B, 70B      & Llama 2 Community  & \cite{chen2023meditron} \\ 
MentaLLama            & Sep 2023                & Mental Health Classification Datasets                                                                      & Llama-2-in              & 13B        & MIT              & \cite{yang2024mentallama} \\ \hline
\multicolumn{6}{l}{\textbf{PROPRIETARY MODELS}} \\
GPT4o            & Sep 2024                & Unknown                                                                      & N/A             & Unknown, reported to be > 200B        &   Proprietary            & \cite{achiam2023gpt} \\ \hline
\end{tabularx}
\caption{Comprehensive information on all LLMs tested against CounselingBench.}
\label{tab:a3}
\end{table*}

\begin{table*}[ht!]
\centering
\begin{tabular}{p{0.9\textwidth}}
\hline
You are a helpful, respectful, honest, and knowledgeable student studying to become a licensed therapist. You must answer a series of multiple-choice questions provided by the user from a US mental health counselor licensing exam. Based on the question text and the context provided, you must answer with either "A", "B", "C", or "D". \\

\#\#\# USER: **Question**: \newline
The patient demographic is as follows: [PATIENT DEMOGRAPHICS] \newline
[PRESENTING PROBLEM] \newline
[MENTAL STATUS EXAMINATION] \newline
[OTHER CONTEXTS] \newline
Given the context above and your expert-level knowledge of mental health counseling, please answer the following question: [QUESTION] \newline
[CANDIDATE ANSWERS] \\

\#\#\# ASSISTANT: Correct Answer is ( \\

\hline
\end{tabular}
\caption{Template for zero-shot prompting}
\label{tab:a4}
\end{table*}

\begin{table*}[ht!]
\begin{tabular}{p{0.9\textwidth}}
\hline
You are a helpful, respectful, honest, and knowledgeable student who is studying to become a licensed therapist. You must answer a series of multiple-choice questions given by the user from a US mental health counselor licensing exam, based on the question text and the context provided. You must answer with either "A", "B", "C", or "D". \\[1ex]
\#\#\# USER: \textbf{Question:} \\
The patient demographic is as follows: [PATIENT DEMOGRAPHICS] \newline
[PRESENTING PROBLEM] \newline
[MENTAL STATUS EXAMINATION] \newline
[OTHER CONTEXTS] \newline
Given your expert-level knowledge of mental health counseling, please answer the following question: \newline [QUESTION]\newline
[CANDIDATE ANSWERS] \\
\#\#\# ASSISTANT: Correct Answer is ([CORRECT ANSWER]) \\[1ex]
\textit{[Template repeats 2 more times]} \\[1ex]
\#\#\# USER: \textbf{Question:}\textbackslash{}n \\
The patient demographic is as follows: [PATIENT DEMOGRAPHICS] \newline
[PRESENTING PROBLEM] \newline
[MENTAL STATUS EXAMINATION] \newline
[OTHER CONTEXTS] \newline
Given your expert-level knowledge of mental health counseling, please answer the following question: \newline [QUESTION] \newline
[CANDIDATE ANSWERS] \newline
\#\#\# ASSISTANT: Correct Answer is ( \\
\hline
\end{tabular}
\caption{Template for few-shot prompting}
\label{tab:a5}
\end{table*}

\begin{table*}[t]
\begin{tabular}{p{0.9\textwidth}}
\hline
You are a helpful, respectful, honest, and knowledgeable student who is studying to become a licensed therapist. You must answer a series of multiple-choice questions given by the user from a US mental health counselor licensing exam, based on the question text and the context provided. You must answer with either "A", "B", "C", or "D".\\

\textbf{\#\#\# USER: Question:}\newline
The patient demographic is as follows: [PATIENT DEMOGRAPHICS] \newline
[PRESENTING PROBLEM] \newline
[MENTAL STATUS EXAMINATION] \newline
[OTHER CONTEXTS] \newline
Given your expert-level knowledge of mental health counseling, please answer the following question by carefully and thoroughly reason step-by-step, leveraging relevant facts from the question context and expert-level counseling knowledge, then clearly indicate your answer with "Therefore, the correct answer is (A)", "Therefore, the correct answer is (B)", "Therefore, the correct answer is (C)" or "Therefore, the correct answer is (D)" at the end of your answer: \newline[QUESTION]\newline
[CANDIDATE ANSWERS]\\
\textbf{\#\#\# ASSISTANT:} [EXPERT-WRITTEN REASONING CHAIN]. Therefore, the correct answer is ([CORRECT ANSWER])\\\\

\textit{[Template above repeats 2 more times]} \\[1ex]

\textbf{\#\#\# USER: Question:}\textbackslash{}n\\
The patient demographic is as follows: [PATIENT DEMOGRAPHICS] \newline
[PRESENTING PROBLEM] \newline
[MENTAL STATUS EXAMINATION] \newline
[OTHER CONTEXTS] \newline
Given your expert-level knowledge of mental health counseling, please answer the following question by carefully and thoroughly reason step-by-step, leveraging relevant facts from the question context and expert-level counseling knowledge, then clearly indicate your answer with "Therefore, the correct answer is (A)", "Therefore, the correct answer is (B)", "Therefore, the correct answer is (C)" or "Therefore, the correct answer is (D)" at the end of your answer: \newline[QUESTION]\newline
[CANDIDATE ANSWERS]\newline
\textbf{\#\#\# ASSISTANT:} \\
\hline
\end{tabular}
\caption{Template for few-shot, chain-of-thought prompting}
\label{tab:a6}
\end{table*}

\begin{table*}[]
\centering
\begin{tabular}{lllllll}
\hline
setting                                      & Overall & IA\&D & TP & CS\&I & PP\&E & CCA \\
\hline
med42-70b                 & 0.558             & 0.642                                       & 0.516                        & 0.531                                         & 0.573                                      & 0.386                                \\
\rowcolor{LightCyan2} 
meditron-70b             & 0.604             & 0.629                                       & 0.631                        & 0.565                                         & 0.581                                      & 0.568                                \\
medalpaca-13b           & 0.103             & 0.148                                       & 0.051                        & 0.079                                         & 0.167                                      & 0.110                                \\
\rowcolor{LightCyan2} 
Llama-2-70b-in     & 0.626             & 0.653                                       & 0.636                        & 0.625                                         & 0.607                                      & 0.437                                \\
BioMedGPT-7B      & 0.442             & 0.501                                       & 0.499                        & 0.357                                         & 0.248                                      & 0.409                                \\
\rowcolor{LightCyan2} 
Llama-3-70B-in & 0.733             & 0.742                                       & 0.732                        & 0.739                                         & 0.703                                      & 0.700                                \\
Llama-2-13b-in    & 0.533             & 0.566                                       & 0.569                        & 0.511                                         & 0.532                                      & 0.484                                \\
\rowcolor{LightCyan2} 
gpt4o       & 0.731             & 0.742                                       & 0.737                        & 0.729                                         & 0.713                                      & 0.609                                \\
Llama3-Med42-70B          & 0.706             & 0.724                                       & 0.706                        & 0.698                                         & 0.683                                      & 0.608                                \\
\rowcolor{LightCyan2} 
Llama-3-8B-in  & 0.67              & 0.540                                       & 0.748                        & 0.809                                         & 0.499                                      & 0.324                                \\

OpenBioLLM-8B      & 0.574             & 0.36                                        & 0.746                        & 0.722                                         & 0.370                                      & 0.554                                \\
\rowcolor{LightCyan2} 
Asclepius-13B         & 0.308             & 0.292                                       & 0.310                        & 0.314                                         & 0.345                                      & 0.35                                 \\
medalpaca-7b            & 0.154             & 0.224                                       & 0.053                        & 0.169                                         & 0.215                                      & 0.163                                \\
\rowcolor{LightCyan2} 
Asclepius-Llama3-8B   & 0.425             & 0.494                                       & 0.168                        & 0.586                                         & 0.418                                      & 0.326                                \\
Llama3-Med42-8B         & 0.638             & 0.632                                       & 0.66                         & 0.651                                         & 0.610                                      & 0.563                                \\
\rowcolor{LightCyan2} 
Llama-2-7b           & 0.327             & 0.336                                       & 0.308                        & 0.333                                         & 0.327                                      & 0.474                                \\
Llama-2-7b-in     & 0.445             & 0.441                                       & 0.505                        & 0.447                                         & 0.440                                      & 0.341                                \\
\rowcolor{LightCyan2} 
Llama-2-13b         & 0.420             & 0.420                                       & 0.439                        & 0.415                                         & 0.358                                      & 0.660 \\
\hline
\end{tabular}
\caption{Few-shot accuracy of tested LLMs on CounselingBench, segmented across different counseling competencies}
\label{tab:a7}
\end{table*}

\begin{table*}[]
\centering
\begin{tabular}{lllllll}
\hline
setting                                    & Overall & IA\&D & TP & CS\&I & PP\&E & CCA \\
\hline
medalpaca-7b             & 0.372             & 0.375                                       & 0.413                        & 0.347                                         & 0.380                                      & 0.303                                \\
\rowcolor{LightCyan2} 
Llama3-Med42-8B         & 0.654             & 0.647                                       & 0.657                        & 0.653                                         & 0.614                                      & 0.611                                \\
med42-70b                 & 0.687             & 0.684                                       & 0.701                        & 0.691                                         & 0.663                                      & 0.654                                \\
\rowcolor{LightCyan2} 
ClinicalCamel-70B         & 0.702             & 0.730                                       & 0.687                        & 0.689                                         & 0.693                                      & 0.559                                \\
Llama3-OpenBioLLM-8B      & 0.603             & 0.599                                       & 0.634                        & 0.600                                         & 0.583                                      & 0.442                                \\
\rowcolor{LightCyan2} 
Llama-2-70b           & 0.642             & 0.659                                       & 0.645                        & 0.643                                         & 0.628                                      & 0.604                                \\
Llama-2-7b\             & 0.370             & 0.371                                       & 0.367                        & 0.373                                         & 0.351                                      & 0.306                                \\
\rowcolor{LightCyan2} 
medalpaca-13b             & 0.260             & 0.264                                       & 0.265                        & 0.277                                         & 0.275                                      & 0.175                                \\
llama-2-7b-in        & 0.456             & 0.448                                       & 0.494                        & 0.443                                         & 0.460                                      & 0.256                                \\
\rowcolor{LightCyan2} 
meditron-70b              & 0.632             & 0.650                                       & 0.637                        & 0.626                                         & 0.631                                      & 0.565                                \\
Llama-2-13b            & 0.475             & 0.462                                       & 0.506                        & 0.451                                         & 0.478                                      & 0.388                                \\
\rowcolor{LightCyan2} 
Llama-3-8B-in  & 0.654             & 0.664                                       & 0.614                        & 0.671                                         & 0.645                                      & 0.614                                \\
Asclepius-Llama3-8B       & 0.385             & 0.432                                       & 0.358                        & 0.383                                         & 0.343                                      & 0.390                                \\
\rowcolor{LightCyan2} 
Llama-2-70b-in       & 0.635             & 0.650                                       & 0.628                        & 0.631                                         & 0.634                                      & 0.485                                \\
MentaLLaMA-13B-in       & 0.488             & 0.496                                       & 0.543                        & 0.482                                         & 0.473                                      & 0.519                                \\
\rowcolor{LightCyan2} 
Asclepius-13B             & 0.380             & 0.357                                       & 0.3666                       & 0.381                                         & 0.407                                      & 0.452                                \\
Llama3-Med42-70B          & 0.701             & 0.723                                       & 0.710                        & 0.709                                         & 0.681                                      & 0.659                                \\
\rowcolor{LightCyan2} 
Asclepius-7B              & 0.303             & 0.283                                       & 0.289                        & 0.295                                         & 0.328                                      & 0.385                                \\
Llama3-OpenBioLLM-70B     & 0.734             & 0.745                                       & 0.715                        & 0.727                                         & 0.715                                      & 0.701                                \\
\rowcolor{LightCyan2} 
BioMedGPT-7B           & 0.431             & 0.448                                       & 0.454                        & 0.427                                         & 0.425                                      & 0.479                               \\
meditron-7b\            & 0.307             & 0.322                                       & 0.299                        & 0.307                                         & 0.326                                      & 0.221                                \\
\rowcolor{LightCyan2} 
Llama-2-13b-in    & 0.542             & 0.551                                       & 0.584                        & 0.518                                         & 0.537                                      & 0.479                                \\
Llama-3-70B-in & 0.733             & 0.732                                       & 0.718                        & 0.751                                         & 0.693                                      & 0.742 \\   
\hline
\end{tabular}
\caption{Few-shot with self-consistency accuracy of tested LLMs on CounselingBench, segmented across different counseling competencies}
\label{tab:a8}
\end{table*}

\begin{table*}
\begin{tabularx}{0.4\textwidth}{p{0.15\textwidth}p{0.35\textwidth}p{0.4\textwidth}}
\hline
\textbf{Metric} & \textbf{Definition} & \textbf{Formula} \\
\hline
Faithfulness (faith) & Measures if model misinterpreted problem statement, or if reasoning chain is vague, irrelevant or misuses information. Step-level score based on alignment from hypothesis steps to source sentences; calculated as mean reasoning alignment score over steps. & \(\frac{1}{N}\sum_{i=1}^N[1 + max_{j=1}^T cos(h_i, s_j)] / 2\) \\
\hline
Informativeness - Step (\(info_{stp}\)) & Measures how well source information is used in reasoning steps. Gives higher score to well-grounded reasoning steps. Identifies degree of source information covered by generated hypothesis. Lower score indicates steps unrelated to source or missing context information. & \(\frac{1}{T} \sum_{t=1}^T r_{align}(s_t \to h) + (1/N)\sum_{i=1}^N r_{align}(h_i \to s)] / 2\)\\
\hline
Informativeness - Chain (\(info_{chn}\)) & Quantifies agreement between hypothesis chain and source. Embeds reasoning chain and source context as whole, rather than using step-wise embeddings as in *-Step metrics. & \([1 + cos(h, s) / 2]\) \\
\hline
Missing Step (mis.) & Identifies steps missing from hypothesis but required for problem-solving. Examines alignment between reference and hypothesis, checking each reference step for similar hypothesis step. & \(min_{i=1...K}(r_{align}(h_i \to r)\)\\
\hline
Reasoning Alignment (al.) & Evaluates hypothesis chain correctness by comparing overlap between hypothesis and reference through measuring reasoning alignment. & \((1/N) \sum_{i=1}^N r_{align}(h_i \to r)\)\\
\hline
Self-Consistency (cons.) & Measures logical entailment errors within reasoning steps. Penalizes chains where step pairs likely contradict each other. & \(1 - max_{i=2...N} max_{j < i}p_{contr}(h_i, h_j)\)\\
\hline
Repetition (rep.) & Identifies repeated or paraphrased steps using token alignment scores between all steps in hypothesis chain. & \(1 - max_{i=2...N}max_{j=1...i-1}[cos(h_i, h_j)]) / 2\)\\
\hline
Grammar (gmr.) & Probability of grammatical acceptability of each step, averaged over all steps & \((1/N) \sum_{i=0}^N p_gram(h_i)\)\\
\hline
\end{tabularx}
\caption{Detailed information and formula on selected subset of ROSCOE reasoning metrics, as defined in \cite{golovneva2022roscoe}}
\label{tab:a9}
\end{table*}

\begin{table*}[t]
\setlength{\tabcolsep}{2pt}
\renewcommand{\arraystretch}{1.1}
\begin{tabularx}{\textwidth}{lX}
\hline
\textbf{Error Category} & \textbf{Definition} \\ \hline
Knowledge Error & The model's reasoning cites domain knowledge that is either incorrect or irrelevant to the current question \\ \hline
Context Error & The model either fails to recall details from the context that is crucial to answer the question, or it recalls the incorrect details from the context \\ \hline
Logical Error & The model's reasoning steps misaligns with the logic used in the corresponding expert-written reasoning \\ \hline
\end{tabularx}

\vspace{1mm}

\begin{tabularx}{\textwidth}{p{\textwidth}}
\textbf{Examples and Explanations} \\ \hline
\textbf{Knowledge Error Example:} \\
\textit{Question:} During the session, the client states, "I don't see how things will improve because things have been bad for so long." Which cognitive distortion is this? \\
\textit{Choices:} (A) Catastrophizing, (B) Black-and-white thinking, (C) Minimizing, (D) Generalization \\
\textit{LLM Reasoning:} The client's statement reflects catastrophizing, which involves exaggerating negative consequences. Therefore, (A) is correct. \\
\textit{Error Explanation:} The LLM incorrectly defined catastrophizing as "exaggerating negative consequences" when it actually means "thinking the worst-case scenario will happen" \\ \hline

\textbf{Context Error Example:} \\
\textit{Question:} What can you do to build comfort with a client when talking about interests hasn't worked? \\
\textit{Choices:} (A) Explain therapy benefits, (B) Play a board game, (C) Use silence, (D) Use immediacy skill \\
\textit{LLM Reasoning:} Using immediacy skill (D) to address the client's feelings about the session can help create trust. \\
\textit{Error Explanation:} The LLM failed to recall that immediacy had already been tried in previous sessions, making another comfort-building strategy necessary \\ \hline

\textbf{Logical Error Example:} \\
\textit{Question:} Which planning objective would best support Taylor's stated goal? \\
\textit{Choices:} \\
(A) Identify three anxiety triggers \\
(B) Manage emotions during transitions \\
(C) Learn three coping methods in six months \\
(D) Apply one coping method successfully \\
\textit{LLM Reasoning:} (B) is best as it's SMART (specific, measurable, achievable, relevant, time-bound). \\
\textit{Error Explanation:} The LLM asserts (B) is SMART without justification, when it actually lacks specificity and timeline \\ \hline
\end{tabularx}
\caption{Definition of 3 main categories of reasoning errors, along with random examples for each category}
\label{tab:a10}
\end{table*}

\begin{table*}[t]
\begin{tabularx}{\textwidth}{lX}
\hline
\multicolumn{2}{p{0.97\textwidth}}{\textbf{Question:} When discussing treatment issues, you ask, "How does drinking align or not align with your goal of improving relationships with your family?" What is the value of posing this question?} \\
\hline
\textbf{Llama3-70B-in} & The question helps create a discrepancy between the client's actions (drinking) and her personal values (improving relationships with her family), which can motivate her to make changes. By exploring this discrepancy, the counselor can help the client see how her behavior is not aligning with her goals and values, which can increase her motivation to address her alcohol use. Therefore, the correct answer is (B) \\
\hline
\textbf{OpenBioLLM70B} & This question helps create a discrepancy between the client's actions (drinking) and personal values (improving relationships with her family). By highlighting this inconsistency, the counselor can facilitate the client's exploration of the impact of her alcohol use on her relationships and overall well-being. This approach aligns with the principles of motivational interviewing, which often involves helping clients recognize and resolve ambivalence towards change. Therefore, the correct answer is (B) \\
\hline
\textbf{Llama3-Med42-70B} & The question posed, "How does drinking align or not align with your goal of improving relationships with your family?" is valuable because it helps create a discrepancy between the client's actions (heavy alcohol use) and her personal values (improving relationships, especially during this period of significant family upheaval). By highlighting this discrepancy, it can motivate change, as the client begins to see that her alcohol consumption might be counterproductive to achieving her relationship goals. Therefore, the correct answer is (B). \\
\hline
\textbf{GPT4o} & The question posed to the client is designed to help her reflect on how her drinking behavior aligns or conflicts with her personal values and goals, particularly regarding her relationships with her family. This approach is rooted in motivational interviewing, which aims to enhance the client's intrinsic motivation to change by creating a discrepancy between their current behaviors and their broader goals and values. Let's analyze the options: (A): While keeping the focus on the client's alcohol misuse is important, the question specifically aims to explore the alignment between her drinking and her personal goals, rather than just focusing on the misuse itself. (B): This option aligns with the principles of motivational interviewing. By highlighting the discrepancy between the client's actions (drinking) and her personal values (improving relationships with her family), the therapist can help the client recognize the need for change. (C): Providing baseline measures for treatment plan goals is important, but this question is more about creating awareness and motivation rather than establishing baseline measures. (D): Incentive-based interventions used in contingency management involve providing tangible rewards for positive behaviors. This question is more about internal motivation and self-reflection rather than external incentives. Therefore, the correct answer is (B). \\
\hline
\textbf{Meditron-70B} & This question helps create a discrepancy between the client's actions and personal values. Therefore, the correct answer is (B) \\
\hline
\end{tabularx}
\caption{Reasoning chain of top-performing LLMs on an example question from CounselingBench}
\label{tab:a11}
\end{table*}

\begin{table*}[t]
\centering
\footnotesize
\setlength{\tabcolsep}{1.5pt}
\begin{tabular}{lccccccc}
\hline
\textbf{Rationale Comparisons} & \textbf{\begin{tabular}[c]{@{}c@{}}Alig-\\nment\end{tabular}} & \textbf{\begin{tabular}[c]{@{}c@{}}Compre-\\hension\end{tabular}} & \textbf{Recall} & \textbf{Bias} &  \textbf{Harm} & \textbf{\begin{tabular}[c]{@{}c@{}}Irrel-\\evance\end{tabular}} & \textbf{\begin{tabular}[c]{@{}c@{}}Omi-\\ssion\end{tabular}}  \\
\hline
\multicolumn{8}{l}{\textbf{\textit{LLM vs Human}}} \\
Llama3-OpenBioLLM-70B & \textbf{490} & \textbf{762} & 485 & 116 & 370 & 561 & 313 \\
Human & 453 & 738 & \textbf{758} & \textbf{177} & \textbf{582} & \textbf{921} & \textbf{1257} \\
Equal & 666 & 105 & 368 & 1317 & 660 & 129 & 39 \\
\hline
\multicolumn{8}{l}{\textbf{\textit{LLM vs LLM}}} \\
Llama3-OpenBioLLM-70B & 329 & \textbf{687} & 387 & 65 & 237 & 427 & 238 \\
Meta-Llama-3-70B-Instruct & \textbf{332} & 612 & \textbf{521} & \textbf{128} & \textbf{452} & \textbf{973} & \textbf{1311} \\
Equal & 951 & 309 & 704 & 1419 & 923 & 211 & 63 \\
\hline
\end{tabular}
\caption{Number of times a model's rationale is dominant compared by LLM judge across different categories. This was computed by using gpt4o to compare two rationales and decide which performs better in categories, such as medical alignment, reading comprehension, knowledge recall, bias, harm, irrelevance, omission.}
\label{tab:a12}
\end{table*}

\end{document}